Corresponding Author:

Félix Buendía, *Universitat Politècnica de València. Camino de Vera s/n, 46022 Valencia, Spain*

Email: fbuendia@disca.upv.es

# Generation of Reusable Learning Objects from Digital Medical Collections: A Qualitative Analysis

Félix Buendía[1], Joaquín Gayoso-Cabada[2], José-Luis Sierra[2]

[1]*Universitat Politècnica de València. Valencia, Spain*

[2]*Universidad Complutense de Madrid. Madrid, Spain*

## Abstract

Learning Objects represent a widespread approach to structuring instructional materials in a large variety of educational contexts. The main aim of this work consists of analyzing from a qualitative point of view the process of generating reusable learning objects (RLOs) followed by *Clavy*, a tool that can be used to retrieve data from multiple medical knowledge sources and reconfigure such sources in diverse multimedia-based structures and organizations. From these organizations, *Clavy* is able to generate learning objects which can be adapted to various instructional healthcare scenarios with several types of user profiles and distinct learning requirements. Moreover, *Clavy* provides the capability of exporting these learning objects through educational standard specifications, which improves their reusability features. The analysis insights highlight the importance of having a tool able to transfer knowledge from the available digital medical collections to learning objects that can be easily accessed by medical students and healthcare practitioners through the most popular e-learning platforms.

## Introduction

Learning Objects represent a widespread approach to structuring instructional materials in a large variety of educational contexts from primary school (K-12 levels) to graduate studies and staff training. Medical education is not an exception and this kind of learning resources are used when teaching undergraduate medicine courses or training healthcare professionals. The current work addresses an analysis of how learning objects can be generated in the context of medical education from existing digital collections by using a tool called *Clavy*[1]. The analysis takes into account the technical features that make learning objects to be re-used according to the educational scenario in which they can be deployed. The proposed analysis is performed from a qualitative point of view, which examines those technical features when accessing, retrieving and processing resources in digital medical collections by means of open computing interfaces.

In this sense, there are multiple sources of medical knowledge which are available through the Internet although they are usually very heterogeneous and require a specific processing or curation method to generate Reusable Learning Objects (RLOs). Thus, the main aim of the paper consists of analyzing the process of generating learning objects from these existing digital collections, which can be potentially re-used in several medical education scenarios. This process demands the creation of tools that, like *Clavy*, can support the generation of such RLOs while taking into account the heterogeneity of available medical knowledge sources, the diversity of the multimedia materials and contents these sources are composed of, or the variety of standards that can be deployed in the e-learning platforms where these instructional contents are used.

*Clavy* allows users with different profiles to generate RLOs according to their specific needs and instructional requirements; for example, in e-learning-based online courses that combine medical case assets with interactive quizzes or forum activities. For this purpose, *Clavy* is able to retrieve multiple types of medical knowledge information such as research articles, clinical reports, or imaging databases that integrate various multimedia formats from plain or structured text to complex images and video recordings. This wide diversity of medical knowledge presents a daunting task in the generation of learning objects to be deployed in courses addressed to different user profiles (e.g. undergraduate medical students or resident physicians-in-training). The capability of *Clavy* to reconfigure these diverse knowledge resources into several structures and organizations, from small or medium-size course units to institutional training content, facilitates their reusability in many medical education contexts. Moreover, the possibility provided by *Clavy* to export several instructional contents in the form of standard specifications[2] such as IMS CP (Content Package) or SCORM improves such reusability.

Given that the medical education context covers a wide range of topics, we have focused on the radiology area in order to analyze how the related information can be retrieved, organized and reused for different learning purposes. Radiology teaching files have served for such purposes for a very long time[3]. Smith and Castillo[4] described the Internet-accessible radiology teaching file especially addressed to emergency radiologists. Zeidel et al[5], also presented the concept of a web-based radiology teaching file as a mechanism used by medical students, residents, physicians, and researchers to access Web information associated with clinical studies in this area. D'Allesandro[6] commented on the role of the RadiologyEducation[7] website as a digital library of educational radiology resources, and related initiatives such as Radiopaedia[8] or Radrounds[9] offer case catalogs showing how patients are imaged and diagnosed. In parallel, the emergence of the PACS (Picture Archiving and Communication System) tools provides a technology to store radiology images, thus enabling convenient access to them from multiple modalities and devices[10]. These PACS items and other radiographic images have been brought to generate digital teaching files and imaging repositories[11] and along with the DICOM standard have been used to automatically create teaching file cases[12]. Some of these educational resources are organized into specialized repositories as a MIRC[13] (Medical Imaging Resource Center) for storing teaching files but their access as reusable learning objects remains pending. In summary, there is an extraordinary number of educational resources available, just in the radiology area and its related collections, but the greater challenge is how to convert them into learning objects with suitable features to be reused in multiple kinds of learning or training medical scenarios.

The analysis of the capabilities of the *Clavy* tool to generate RLOs from existing medical digital collections is based on a qualitative point of view, and it uses access to the *MedPix* (https://medpix.nlm.nih.gov) online clinical case database as a comprehensive case study. This case study makes it possible to examine the different *Clavy* capabilities such as retrieving heterogeneous multimedia information sources from this database, organizing and structuring the content elements gathered and supporting the RLO generation process in general. The remainder of the work is structured as follows. We start by reviewing some related works in the generation of RLO in several medical education contexts and more specifically, from the radiology perspective. Then, we provide some insights into *Clavy's* mechanisms used to access medical collections, retrieve and organize their digital contents, and export them in standard formats. Next, we address the qualitative analysis of the *Clavy* approach to generating reusable learning objects, using *MedPix* as a significant case study. It is followed by a discussion on said process of analysis and the criteria applied. We finish with some conclusions and lines of further work.

## Related works

Learning objects have been widely used in medical education as they are considered a "grouping of instructional materials structured to meet a specified educational objective"[14]. This broad definition allows instructors to create a huge variety of materials under the "learning object" umbrella but their development requires a minimum set of features to formally characterize them in terms of granularity, self-content, reusability, aggregation or interoperability[15]. However, drawing such features during the generation of learning objects is a challenging task within a medical educational context. A particular issue is what constitutes an RLO and how to make these objects reusable. The student perspective on this kind of learning objects was described by Blake[16] who commented on some of the barriers to their use. The design and analysis of RLOs in healthcare meta-analysis learning was also described[17]. Windle and Wharrad[18] introduced some reflections on the characteristics of RLOs in healthcare education; for example, on the use of context-neutral materials as a way to improve reusable features. They also mentioned that "reusability also requires that a resource is platform-independent" and the need to apply non-restrictive licensing models. An interesting aspect included in Windle and Wharrad's work was related to the development processes which were involved in the generation of RLOs, their granularity level (e.g. 15 minutes of scheduled learning activity) and the role of communities of practices in these generation processes.

Evans[19] commented on how the development and evaluation of RLOs could enhance the learning experience of international healthcare students. This initiative was launched in a specific educational context related to nursing and referred to two examples of available RLOs in this context. Again, the multiple possibilities when creating this kind of resources make it difficult to have strict rules that guide such a generation process. Instead, each medical discipline seems to adapt this concept of learning object to its own peculiarities, such as the case of testing a Learning Virtual Object (LVO) in the teaching of radiographic cephalometry[20], the example focused on using 3D anaglyph images made from stereo image pairs in online Anatomy and Physiology courses[21], the development of interactive resources to teach intramuscular medication administration for nursing undergraduates[22], and web-based resources based on mixing multimedia elements like RLOs on the hospital care of people with dementia[23] or addressed to occupational therapy education[24].

This diversity of areas in which it is possible to generate this kind of learning objects makes the need to deploy computing tools that can support their generation process evident. Moreover, from the previous RLO examples there is also evidence of diverse multimedia materials and formats which can be aggregated during their development. This aggregation property is a crucial feature when generating these learning objects since it requires the extraction of such material from multiple medical knowledge sources. In the case of radiology training, there is a high number of information sources, as shown in the previous section. These sources are based on a wide range of multimedia formats and different content characteristics. Even, simple resources such as images or video recordings can have a complex component coming from PACS repositories[25] that store them, standard formats as DICOM to share imaging information objects[26], VR (Virtual Reality) methods for transforming clinical imaging data into learning objects[27] or "virtual lectures" by means of Flash animations[28]. In order to support the generation of resources for radiology education, some initiatives have been launched, such as the aforementioned use of a DICOM service to automatically create teaching file cases from PACS[12] or the development of a radiography Digital Teaching Library (DTL)[29]. However, as Scarsbrook et al[30] commented, some specific characteristics should be evaluated when developing a radiological digital teaching file.

Another important aspect to be addressed regarding the generation of learning objects is the use of standards to improve their interoperability so that they can be used across multiple learning platforms. In this sense, there are several specifications, such as IMS Content Package or SCORM, which are applied to create interactive e-learning courses[31] or are used as references in radiology education[32]. Some of the SCORM-compliant learning initiatives mentioned in the previous reference are RadMoodle or RadSCOPE® (Radiology Shareable Content for Online Presentation and Education). Moreover, an authoring tool was designed to transform DICOM images into the SCORM format[33] and the ILIAS platform (https://www.ilias.de/) was deployed using this format to facilitate the reuse of radiology contents[34]. A Web application called USRC (University of Saskatchewan Radiology Courseware)[35] was created to link teaching MIRC cases to Blackboard contents and, more recently, the *RadEd*[36] tool provides a web-based teaching framework that combines the generation of case-based exercises and the integration of image interaction. This interest about "content creation" and its associated challenges is one of the main contributions by *RadEd* and it is highly related to the approach proposed in the current work.

## The *Clavy* approach in RLO generation

This section describes the *Clavy* approach to generating RLOs from existing medical collections. This approach introduces a versatile mechanism to retrieve information from these collections, reconfigure it according to diverse structures, and, finally, generate this kind of learning objects in several educational specification formats.

*Clavy* is able to obtain pieces of content that are usually poorly structured in external sources and arrange them in highly specialized digital collections. For this purpose, the *Clavy* tool supports a three-step workflow in the creation of such collections from heterogeneous external digital resources:

- The first step enables the importation of resources coming from external knowledge sources. *Clavy* allows the aggregation of the content of multiple collections by using different general-purpose plug-ins to import data from standard platforms and formats such as XML documents, JSON files or relational databases, as well as more complex plug-ins targeted to specific collections that have to be accessed through a specific API (Application Programming Interface). This feature lets healthcare experts retrieve pieces of content that are usually poorly structured in external digital medical collections and arrange them in a versatile and powerful organization.

- The second step consists of the curation of the resources retrieved. When a collection is imported in *Clavy*, experts can access the collection and edit it in its entirety to enrich it or adapt its structure to specific learning scenarios. Therefore, healthcare experts can curate or reconfigure all the information items imported, thus ensuring a coherent and unified structure and reorganizing the collection to satisfy the needs of final users.

- The third step is concerned with the exportation of the already curated and/or reconfigured collection. Indeed, when the experts consider that the collection is ready, *Clavy* provides a second kind of plug-in to export the complete collection, or part of its content elements to third-party systems like, for example, a SCORM object.

The workflow presented is based on several items that are part of a *Clavy* collection:

- A set of digital resources, which include both local files and external resources represented by Uniform Resource Locators (URLs)

- A set of documents, which represent logical aggregations and organizations of resources in terms of hierarchies of element-value pairs.

- A collection schema, which describes the hierarchical organization of content elements in the collection's documents.

Every document in *Clavy* shares a common structure described by its schema that can be reconfigured by editing said schema. In this way, it is possible to rename element types, remove useless or non-relevant elements, merge two semantically equivalent element types, and, change the hierarchical organization of elements. Figure 1 shows an example of screenshot that displays the *Clavy* schema editor applied to a *MedPix* sample case. In this example, there are three elements in the collection hierarchy, representing the *Cases* information which contains *Patient* data, the *Findings* of the case, or the associated *Discussion*, a *Topics* item and a *Quiz* element. These last two elements are represented in a folded format hiding their inner structure. In this schema view, some of its elements (displayed in red) are atomic and oriented to delimiting descriptive texts. Others, such as *Cases* or *Diagnosis* have an additional hierarchical structure associated which can also be edited in a user-friendly way.

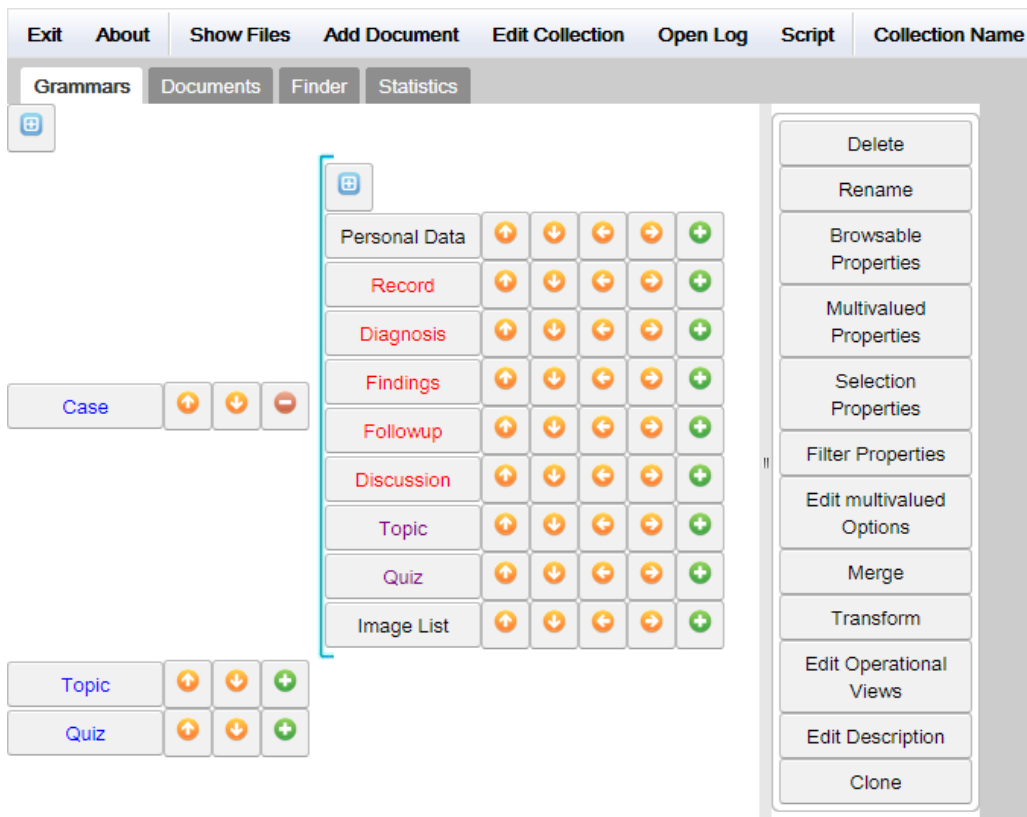

Figure 1. Schema editor in Clavy.

*Clavy* can be used to build reconfigurable content repositories from a diversity of information sources. This feature enables the selection of certain knowledge items according to a specific need, for example, an educational purpose in a healthcare context, and changing such a selection in a dynamic way. Moreover, link elements make it possible to include non-hierarchical relationships in the resulting collection structure, which allows users to enrich the knowledge to be represented with references to internal annotations or external resources. The global process is guided through the following steps in the case of building a content repository with a potential educational purpose:

- First, instructors discover and import digital resources from different sources with educational value suitable to be transformed into learning objects. During this starting step, basic content elements can be collected and also be aggregated by means of hierarchical or non-hierarchical structures. The result in *Clavy* is an initial metadata schema of the content repository.

- Then, instructors can align the organization of the different content items gathered in the initial metadata schema to get an alternative organization adapted to their own educational requirements. They can also use the document editor to curate some elements by editing their content or adding missed instructional information (e.g. some required learning goals or related assessment strategies). Figure 2 shows an example screenshot displaying the *Clavy* document editor that enables information about the *Case* elements to be changed or inserted; for example, its *Diagnosis* components, the *Findings* description or the related *Discussion*.

- Finally, the repository of content objects can be exported using standard specifications such as IMS CP or SCORM, after deciding the final features of such objects; for example, if they include interactive quizzes or links to image annotations and other multimedia items.

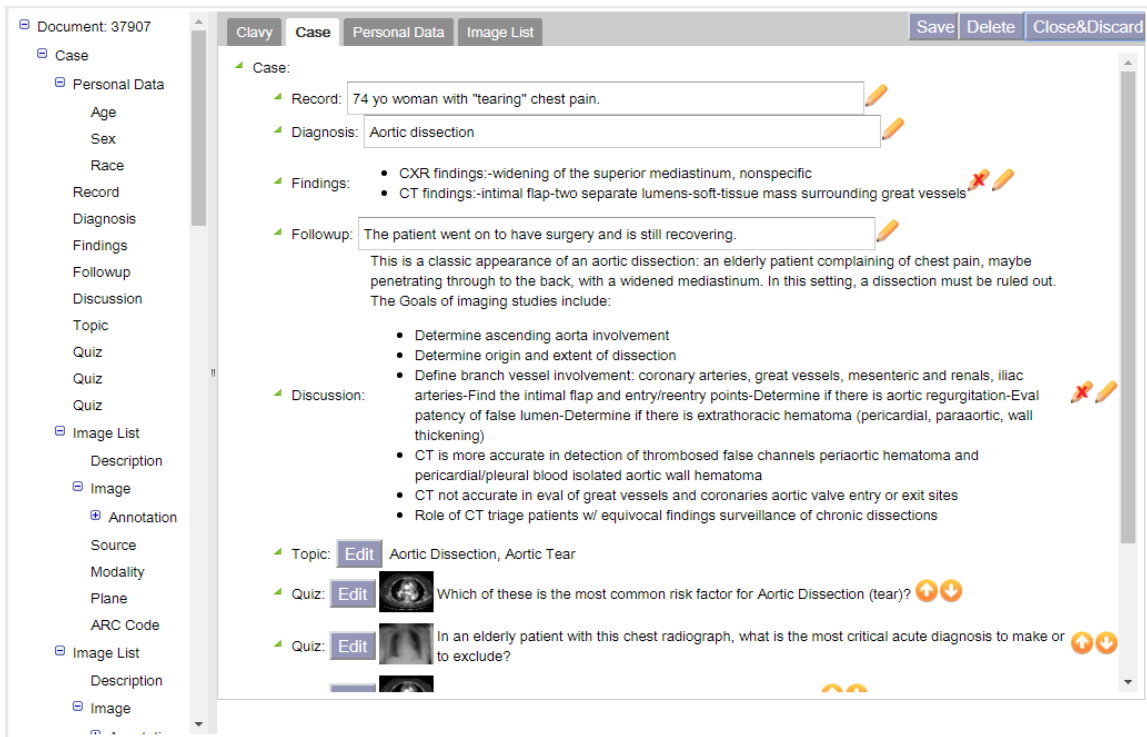

Figure 2. Document editor in Clavy.

A practical example of the *Clavy* approach is described in the next section, which shows how this tool can be deployed to generate learning objects that can be reused in different contexts.

# Analysis of the generation approach

In this section, a sort of qualitative analysis is applied to review the proposed *Clavy* generation approach and those RLOs that can be obtained from this approach. This analysis emphasizes the inspection of the technical aspects that characterize the generation of such reusable learning objects by using the *Clavy* tool. These aspects are summarized in Table 1 and are examined as analysis criteria through the current section. An example of digital medical collection based on a subset of cases collected from the *MedPix* database is used to show these criteria and their application on the *Clavy* content items. These content items include the elements that define the case and topic information along with the questionnaires associated with such *Medpix* cases.

Table 1. Summary of analysis criteria for the Clavy generation approach.

| Criterion | Description |
|---|---|
| *Importation* | Ability to retrieve multiple types of information sources, including different multimedia formats by means of plug-ins. |
| *Granularity* | Readiness to shape the size of learning objects according to user profile or instructional requirements. |
| *Adaptability* | Possibility to build different versions of the same resource (e.g. a clinical case) depending on the required learning goals. |
| *Aggregation* | Ability to integrate several types of resources, either external (URLs) or internal (Annotations). |
| *Interoperability* | Generation of types of resources according to standard specifications such as IMS CP or SCORM. |
| *Personalization* | Ability to offer individual user adaptations; for example, allowing users to select the level of description detail required. |
| *Interactivity* | Capability of learning objects to interact with the user by means of instructional mechanisms such as quizzes or activities. |

In this context, the medical domain is characterized by a knowledge corpus spread over a high number of information sources such as research articles, clinical reports, or imaging databases. As mentioned above, one of the main problems dealing with this huge amount of knowledge is its intrinsic heterogeneity. *Clavy* faces this challenge by means of specialized plug-ins able to import the information coming from heterogeneous sources. In the case of *MedPix*, an importation plug-in was implemented to connect with their API and retrieve a subset of clinical cases (around 6700), 4000 topics and 1700 questions, relevant enough to evaluate this *Importation* capability. Such a plug-in is based on REST call operations which enable the clinical case data and topic details to be obtained, along with questions associated to these cases and, additionally, their list of images to be scraped to compose a sample of *Clavy* collection. Figure 3a) shows an excerpt of the initial schema for this collection, which mirrors the basic *MedPix* case structure in *Clavy* terms. This initial schema includes data related to the patient such as sex or age along with more specific information such as the *Findings*, *Diagnosis* or *Discussion* of the case. Once these content elements are stored as a *Clavy* collection, they have to be curated by means of the schema and document editors. Thus, the original schema contained 88 elements, many of which were not excessively interesting from an educational point of view (for example, some technical details of the image list). Moreover, information concerning the patient in a clinical case could be grouped behind a *Personal Data* structural element or a specific *Image* selected from the available list. Figure 3b) shows the resulting *Clavy* schema after these reconfigurations. The 88 initial elements were reduced to 33 items, many of them grouped in alternative structures.

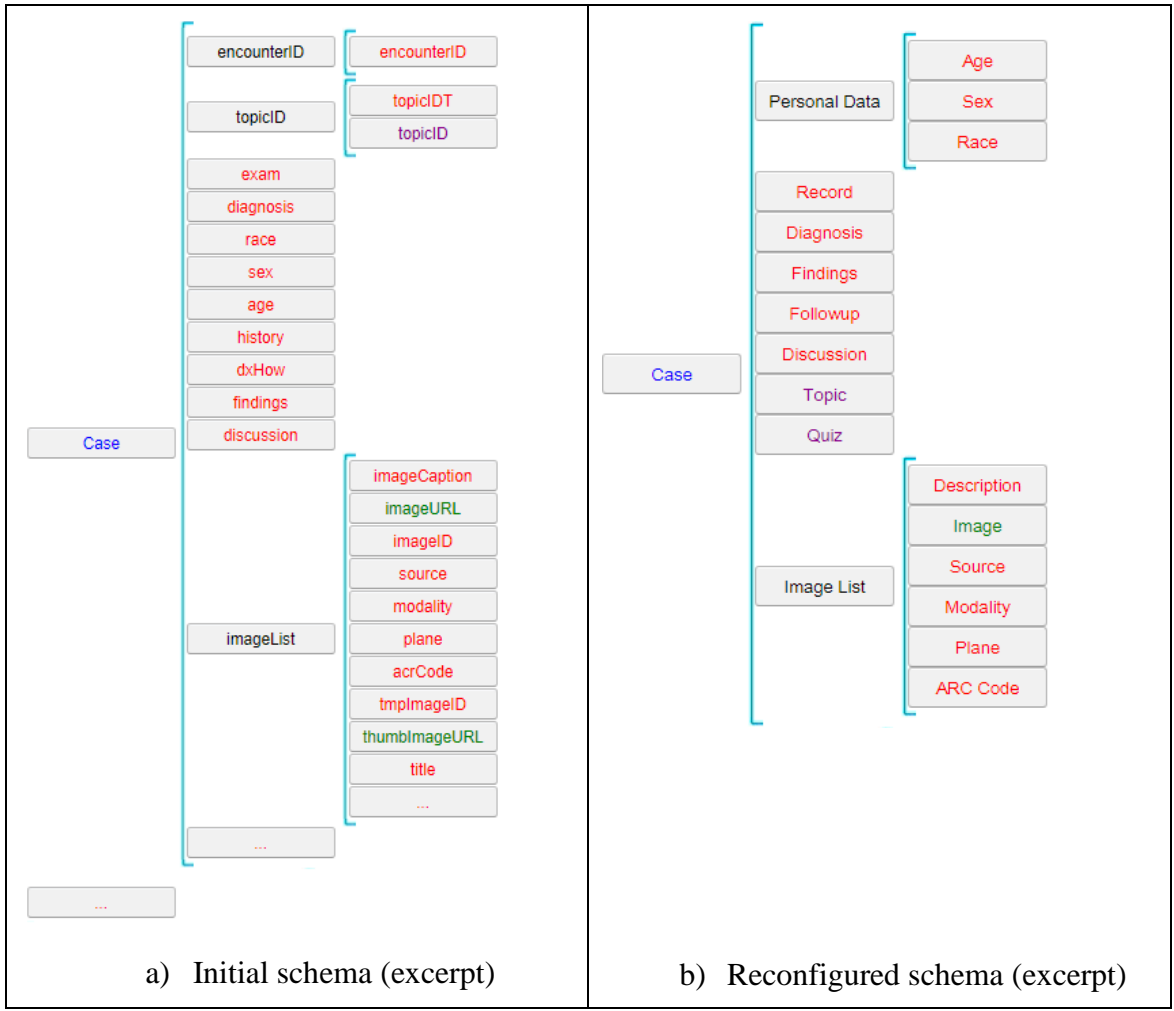

a)  Initial schema (excerpt)          b)  Reconfigured schema (excerpt)

Figure 3. Schema reconfiguration during the Clavy curation process.

This reconfiguration procedure allows practitioners to choose the most suitable object size according to their educational needs and complement such tailored information with additional data that make up the required learning object or a specific package thereof. Such a procedure is crucial when analyzing aspects such as the *Granularity* of the learning objects to be generated or their *Adaptability* to different educational purposes.

The definition of smaller or larger learning objects can be used to support a fine-grain instructional process (for example, oriented towards undergraduate students in 1st year courses who learn basic medical concepts) versus a more detailed information model addressed to representing complex clinical cases to be studied by resident physicians. Moreover, the same clinical case can be adapted to show basic case information to be examined or to integrate interactive quizzes to assess user understanding of the case considered. That is an example of *Clavy's* potential to improve the reusability of those learning objects generated as outcomes, which can be further shared in educational repositories and e-learning platforms.

Another aspect to be inspected is *Clavy's* ability to integrate different types of contents that could be part of the learning objects produced. These contents can be external, with references to URL resources, or come from the same *Clavy* structure model in form of annotations or links to internal resources (e.g. a list of x-ray images). Figure 4 shows a screenshot with an example of annotation incorporated in the *Clavy* document structure that displays a comment over an injury detected in one of the case images. This *Aggregation* property is very important when adding multimedia items that enrich the information model of the learning object to be generated.

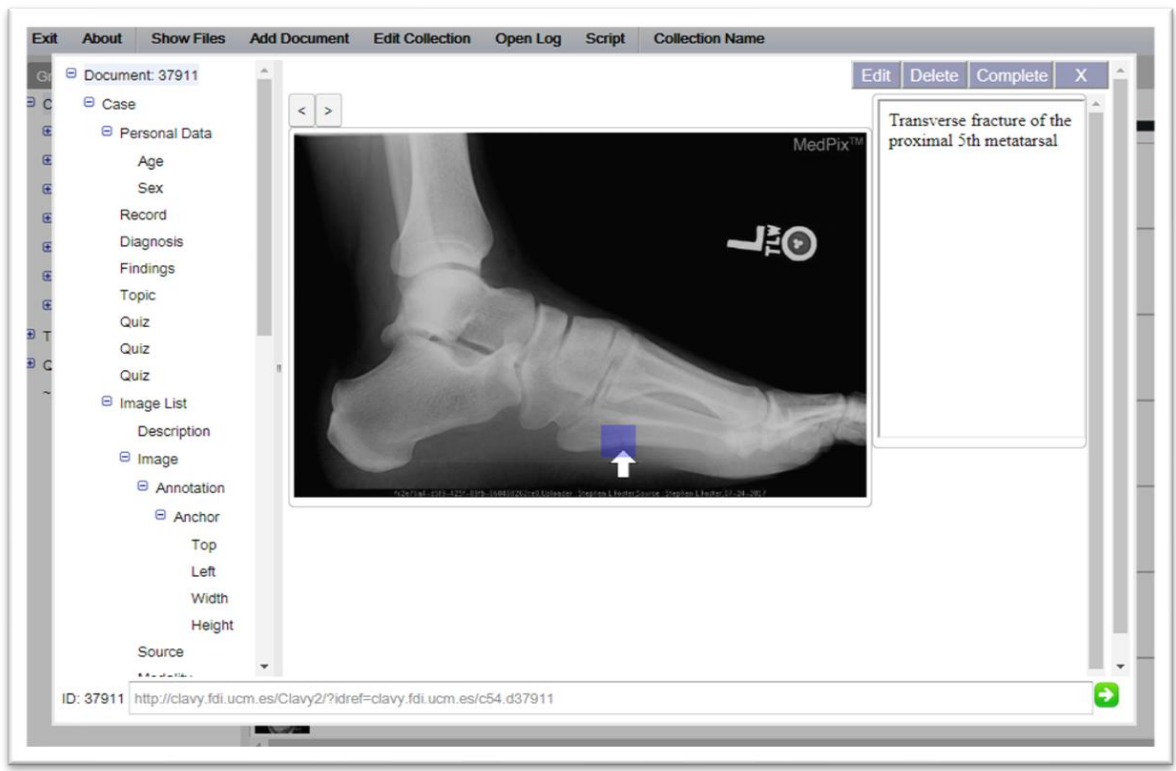

Figure 4. Sample of annotation incorporated in a Clavy document.

The next property to be analyzed is the *Interoperability* provided by *Clavy* when generating learning objects under several standard specifications. At this moment, *Clavy* is able to output collections in the form of IMS Content Packages and SCORM items by using several exportation plug-ins. Figure 5 shows a screenshot that displays part of a SCORM object obtained from a case sample represented through a *Clavy* document. In this example, information from a foot-related case was focused on *Diagnosis* details and a short description of *Findings* to demonstrate the *Personalization* capability of *Clavy* along with the inclusion of *Quiz* items to show the possibilities of interaction.

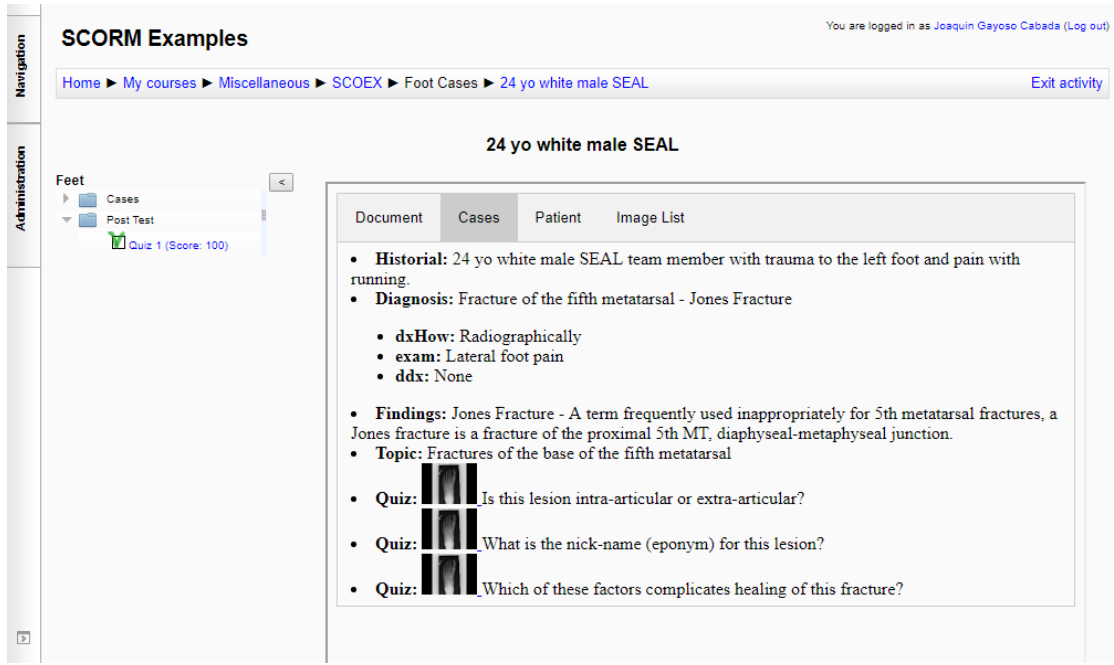

Figure 5. SCORM sample about a foot case coming from the MedPix collection.

In this sense, *Interactivity* is an interesting property to be examined in the case of learning objects that can be enriched with questions to enable their assessment or activities to track user interaction. Figure 6 shows a screenshot that displays an example question attached to the previously mentioned foot case and retrieved from the *MedPix* collection. In this example, the question is part of a SCORM object within a Radiology course offered through a Moodle platform, which shows the *Interactivity* capability that can be introduced in the learning object scope. The question sample is based on an MCQ (Multiple Choice Question) format that asks the user for possible findings about the image displayed and the type of lesion that can be detected using such findings.

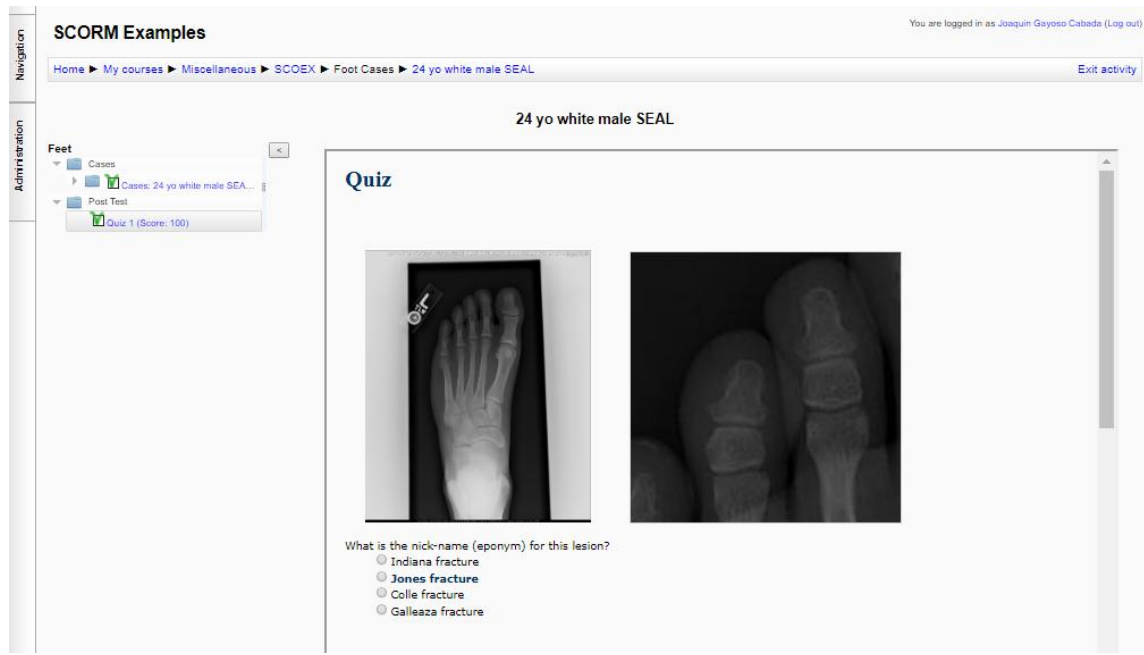

Figure 6. Set of questions attached to a sample of SCORM object.

The technical aspects previously described reveal the several capabilities provided by *Clavy* to generate RLOs. Next section discusses some of these aspects and others that could be addressed in future works.

## Discussion

Previous sections have described an approach to generating RLOs based on the *Clavy* tool and a set of criteria to analyze such an approach and its learning outcomes. Now, it is time to discuss some issues derived from the suitability of the approach proposed and the adequacy of the analysis criteria to examine the RLOs generated.

First, the concept of learning object used through the article has to be viewed from a wide perspective[14] that is open to multiple types of instructional materials and addressed to meeting diverse learning goals. Thus, the concept considered is far from being bound to a specific standard with particular metadata requirements, such as IEEE LOM[2] for example, and the approach proposed fits better with the generation of learning objects whose origin can be found in existing knowledge sources. Generating learning objects is a time-consuming task so we can take advantage of the huge amount of knowledge information in available digital medical collections. Some problems derived from this situation are the high degree of heterogeneity of the aforementioned medical knowledge sources, the diversity of scenarios in which these objects can be implemented or the licensing issues associated to their usage, which were commented by Wharrad and Windle[18]. Regarding the heterogeneity of knowledge sources, one of *Clavy's* main contributions is the possibility of managing it by means of specialized "importation" plug-ins, in order to allow practitioners to exploit the instructional value of those prestigious digital medical collections that are present in the diverse healthcare disciplines.

Dealing with the diversity issue, *Clavy* has proved to be an efficient "collector" of information that can be poorly structured in origin. This capability to organize and reconfigure knowledge is another important contribution that enables the medical information to be selected or adapted according to user requirements either for educational or other purposes. *Clavy* cannot be considered a learning object repository but a tool to organize content items that after a curation process could potentially become learning objects. So, in *Clavy* terms, we are managing content elements that require processing in order to be further deployed in a medical learning context. One of the advantages of *Clavy* is that these content elements can be easily shaped to suit the learning requirements in a specific healthcare educational scenario. In addition, this shaping or curation process leverages *Clavy's* potential to adapt or personalize such elements in a way that improves their reusability. That is a key feature in RLO generation along with the option to export such content elements to standardized formats that make them interoperable in the several e-learning platforms where healthcare courses can be delivered. The future use of LOD (Linked Open Data) formats will improve the reusability and sharing capability of these medical educational resources[37]. Moreover, the possibility to add or infer new knowledge from the originally retrieved *Clavy* collections is another contribution to enhancing the reusability of the learning objects generated. Currently, it is possible in *Clavy* to add annotations over the images that are part of a radiology report used for teaching and to include links to external sources such as bibliographic references to electronic medical libraries or online medical databases but, in the future such annotations could be automatically retrieved from a DICOM-based imaging system, or these bibliography references be extracted through automatic text mining processes in such databases. These features will open new strategies for RLO generation, along with the aggregation of multimedia advances in 3D image processing, virtual reality applications or X-ray simulation tools[38].

Of course, there are also limitations in the *Clavy* application, which are mainly related to the searchability of its collection contents. At this moment, there is no concern to make these collections "searchable" since *Clavy* is far from being a learning repository and becoming an OER (Open Educational Resources) provider for health sciences, like those mentioned by Minter in the UNMC (University of Nebraska Medical Center) blog[39]. Nevertheless, it would be feasible to incorporate some kind of learning object standard model in medical education[40] to tag the objects generated, which would improve *Clavy's* searchability and also enrich their educational contribution. An additional limitation is the fact that *Clavy* cannot be considered a learning platform compared to those examples mentioned in section 2 such as USRC[35] or RadEd[36] but it does include features that could help to generate instructional mechanisms such as interactive questionnaires or educational activities linking multimedia information items. The features provided by *Clavy* when generating standard specifications are in line with the recommendations for their use in the radiology e-learning context[32] and in medical education in general[41]. In the end, *Clavy's* main purpose in the current work is primarily focused on leveraging this generation process from existing medical digital collections and can be considered its greatest contribution[42].

## Conclusions

The current work has presented an approach to generating RLOs based on the deployment of a content management tool called *Clavy*. This approach along with its learning outcomes have been analyzed from a qualitative point of view based on a set of assessment criteria such as the capability to import heterogeneous medical knowledge sources, the adaptation ability in order to shape such imported resources for different learning purposes, the aggregation of various multimedia formats or the interoperability of the learning objects generated that allows them to be used in several e-learning platforms. *Clavy* has proved to be a suitable tool to produce this kind of learning objects in a user-friendly way, enabling the integration of specialized medical collections in the radiology area. Future works include the possibility of extending the scope of medical collections to be processed or testing the RLOs generated in specific medical learning scenarios.

## Acknowledgments

Thanks to Dina Demner Fushman from the *MedPix* and *Open-I* team.

## Fundings

This work was supported by Spanish Research Projects TIN2014-52010-R and TIN2017-88092-R.